\pdfoutput=1
\documentclass[10pt,twocolumn]{article} 
\usepackage{simpleConference_alex}
\usepackage{amssymb}
\usepackage{amsthm}
\usepackage{amsmath}
\usepackage{mathtools}
\usepackage{adjustbox,lipsum}
\newcommand*\samethanks[1][\value{footnote}]{\footnotemark[#1]}

\begin{document}

\title{\textbf{Spiking Neural Networks for event-based action
recognition: A new task to understand their advantage}}

\author{Alex~Vicente-Sola \thanks{Corresponding author: alex.vicente-sola@strath.ac.uk}~~\thanks{Neuromorphic Sensor Signal Processing Lab, Centre for Image and Signal Processing, Electrical and Electronic Engineering, University of Strathclyde, Glasgow, UK.}~,
        Davide~L.~Manna \samethanks~,
        Paul~Kirkland \samethanks~,
        Gaetano~Di~Caterina \samethanks~,
       and Trevor~Bihl
\thanks{ Air Force Research Laboratory, Wright Patterson AFB, OH}
}
\maketitle
\thispagestyle{empty}
\begin{abstract}
Spiking Neural Networks (SNN) are characterised by their unique temporal dynamics, but the properties and advantages of such computations are still not well understood. In order to provide answers, in this work we demonstrate how Spiking neurons can enable temporal feature extraction in feed-forward neural networks without the need for recurrent synapses, and how recurrent SNNs can achieve comparable results to LSTM with a smaller number of parameters. This shows how their bio-inspired computing principles can be successfully exploited beyond energy efficiency gains and evidences their differences with respect to conventional artificial neural networks. These results are obtained through a new task, DVS-Gesture-Chain (DVS-GC), which allows, for the first time, to evaluate the perception of temporal dependencies in a real event-based action recognition dataset. Our study proves how the widely used DVS Gesture benchmark can be solved by networks without temporal feature extraction when its events are accumulated in frames, unlike the new DVS-GC which demands an understanding of the order in which events happen. Furthermore, this setup allowed us to reveal the role of the leakage rate in spiking neurons for temporal processing tasks and demonstrated the benefits of "hard reset" mechanisms. Additionally, we also show how time-dependent weights and normalization can lead to understanding order by means of temporal attention.

Code for the DVS-GC task is available.

\end{abstract}

\section{Introduction}\label{sec:introduction}
Research in neuromorphic computing aims at advancing artificial intelligence (AI) by extracting the computing principles behind biological neural networks. To this end, a large body of work has focused on developing Spiking neural networks (SNN), a closer approximation to real brains \cite{maass1997networks}. From an application point of view, SNN's sparse and asynchronous computations have been demonstrated great gains in energy efficiency when implemented in neuromorphic hardware \cite{davies2021advancing}, thus becoming their major selling point. Still, their differences with respect to conventional ANNs go beyond, as their event-driven temporal dynamics provide an alternative paradigm for temporal processing. The potential advantages of this paradigm have often been overlooked; therefore, a study demonstrating its exploitable properties demands attention.

To provide such demonstration, in this work we will make use of the task of event-based action recognition. This is motivated by the fact that SNNs are naturally suited for the processing of event-based data. These networks are able to integrate input over time and their neurons are activated in an event-based manner, hence their application to event-based data has been a topic of interest \cite{Amir2017ALP, kugele2020efficient,Kirkland2021UnsupervisedSI}. Additionally, given the recent surge in popularity of event-based cameras, research on event-based action recognition is a major priority, making neuromorphic video \cite{Lichtsteiner2006A11, Delbrck2010ActivitydrivenEV} the perfect target.

Despite the ample variety of conventional frame-based action recognition datasets, the options for event-based action recognition are very limited \cite{Lee2014RealTimeGI, Amir2017ALP}, forcing many researchers to resort to artificially generated datasets, which either convert frame sequences to events \cite{Bi2017PIX2NVSPC,Gehrig2020VideoTE} or generate them from simulations \cite{Mueggler2017TheED,Rebecq2018ESIMAO}. Alternatively, those works employing real data from an event camera
\cite{fang2021incorporating, manna2022simple, shrestha2018slayer, xing2020new}
have mainly resorted to IBM's DVS Gesture Dataset \cite{Amir2017ALP}. 

In this work, we prove how solving the action recognition task in the DVS Gesture dataset does not require a network implementing temporal feature extraction. Accumulating events into frames and processing them with an image classifier yields $>$95\% accuracy.

To bypass the limitation of DVS Gesture, we propose DVS-Gesture-Chain (DVS-GC), a new task that can only be solved by those systems capable of perceiving the ordering of events in time. 

Perception of order is a fundamental part of many temporal tasks. Specifically, in action recognition the time dependencies defined by relative order are of critical importance. Often, actions have a sequential nature, where they are composed of a smaller set of sub-actions, and perceiving their ordering is essential to identify the overall action. Further to that, the context of an action and its relationships of causality are also based on the perception of order between actions. Consequently, evaluating this capacity is crucial when designing action recognition systems.

In order to evaluate the aforementioned capacity, DVS-GC leverages the DVS-Gesture data and combines its gestures into chains of gestures, making the chain the actual action class to recognise. 

Using this new task, we show how Spiking Neurons enable spatio-temporal feature extraction without the need for recurrent synapses, demonstrating a form of temporal computation which is different from the one in conventional ANNs and providing an alternative approach to time processing. We analyse the differences in this new computing paradigm with respect to conventional Recurrent Neural Networks (RNN), and further develop the current understanding of it by demonstrating the effects of membrane potential leak and reset mechanism. Specifically, we show how the reset by subtraction approach \cite{6707077} can cause slow adaptation to incoming inputs, translating to what we call a "repetition error". Then we prove how this can be alleviated by voltage leak or by using a reset to zero strategy, leading to improved action recognition accuracy.

Finally, we also explore the role of temporal attention when perceiving order through time-dependent weights and normalization, which implement a type of attention that is hard-coded in time.

\section{Related work}

\subsection{Temporal processing}

When processing temporal sequences of stimuli, a property of cognitive systems that is considered essential for the task is working memory, which holds information from previous events and allows to relate it to those perceived later \cite{diamond2013executive, cowan2008differences}. In the field of neural network engineering, working memory has historically been implemented by recurrent connections, and their memory capabilities have been further enhanced by the use of advanced memory cells such as LSTM \cite{hochreiter1997long} and LMU \cite{voelker2019legendre}. More recently, temporal processing tasks \cite{takase2021lessons, zhou2021informer, wei2021masked}
have also been solved by the increasingly popular Transformer architectures \cite{vaswani2017attention}. When using these networks, temporal events are not presented in a succession as they happen; instead, multiple time-steps are accumulated (or the whole sequence in many cases) and then processed offline by the system. These approaches can be considered to implement working memory outside of the neural network by accumulating stimuli over time and then feeding them to the network together as a single input. Transformers have achieved state of the art accuracy in the majority of temporal tasks, but are limited by their computational and memory complexity, which scale as $O(L^2)$ with the sequence length $L$ or $O(L\,logL)$ in efficient versions such as \cite{kitaev2020reformer}. Hence, research in recurrent architectures is still of interest in order to create lighter systems with dynamic memory management.

Regarding SNNs, the state of the art in temporal tasks is based on RNN architectures. The authors in \cite{bellec2018long} proposed Recurrent SNNs (RSNNs) of Leaky integrate-and-fire (LIF) neurons with neuronal adaptation, a process that reduces the excitability of neurons based on preceding firing activity. Their resulting network is tested in the Sequential MNIST (S-MNIST) and TIMIT tasks. Subsequent work applied LSTM cells to SNN networks, achieving higher performance in S-MNIST \cite{lotfi2020long}.

Still, for the processing of visual event-based datasets such as DHP19 or DVS-Gesture, the state of the art is set by feed-forward SNNs with no recurrency \cite{zheng2020going, fang2021deep, kim2021optimizing}. The remaining question is then whether these feed-forward SNNs implement working memory or, on the contrary, the aforementioned tasks do not require a network with temporal feature extraction. The experiments presented in this work will prove how both statements are true.

\subsection{Event-based datasets}
The event-based sensor market is still in its infancy and its still limited commercial adoption has not allowed to collect large volumes of event-based data. Currently, many of the datasets used in computer vision are artificially created from frame-based data or simulations. N-MNIST, N-Caltech101 \cite{orchard2015converting} and DVS-CIFAR10 \cite{li2017cifar10} are three popular datasets created through screen recordings of the original frame-based data with a neuromorphic camera. Alternatively, frame-based datasets have also been converted into events directly through software \cite{Bi2017PIX2NVSPC, Gehrig2020VideoTE, zhu2021eventgan}. Finally, \cite{Rebecq2018ESIMAO, joubert2021event} provide simulators for the generation of synthetic event data.

Still, the most desirable option for the development of event-based systems is to use data from a real-world acquisition. In the present day, most of the available natively neuromorphic datasets are still simple compared to traditional frame-based ones. For classification tasks we can find: N-CARS \cite{sironi2018hats} a binary classification dataset, ASL-DVS \cite{bi2019graph} a 24 class sign language classification task, DailyActionDVS \cite{ijcai2021p240} a 12 class action recognition task and DVS-Gesture \cite{Amir2017ALP} the widely used 11 class action recognition dataset. Other available datasets are the DHP19 \cite{calabrese2019dhp19} pose estimation dataset, the Gen1 \cite{de2020large} and 1 Mpx \cite{perot2020learning} object detection datasets, and DSEC \cite{Gehrig21ral} for optical flow estimation.

A study on the relevance of neuromorphic datasets for SNN evaluation was presented in \cite{iyer2021neuromorphic}. In this work, it was proven that collapsing all events into a single frame and performing recognition with an image classifier did not affect classification performance in N-MNIST or N-Caltech101, but did decrease it in DVS-Gestures. Our results will show how, when the event integration is done in multiple frames instead of just one, DVS-Gestures can also be solved by a non-temporal image classifier, evidencing its lack of temporal complexity.

Regarding non-classification datasets, the available options are ill-suited for rigorous evaluation of temporal processing. Despite temporal information being exploitable to gain improvements in accuracy, object detection or pose estimation tasks can still be solved by accumulating events in time-windows and performing inference in one frame. In the same way, optical flow estimation can be performed using two frames of accumulated events.

\section{Methodology}
\subsection{DVS Gesture Chain} \label{sec.dvs-gc}


The objective is to define an action recognition task in event-based sequences that requires the perception of temporal dependencies, i.e., relationships where the meaning of an action is contingent to those that happened previously. To create such temporal dependencies, DVS-GC leverages the DVS-Gesture dataset and combines $N$ of its gestures $G=\{g_1, g_2, ...,g_{N}\}$ into chains of gestures $G_{c}=\{(g_i, g_j,...,g_k)\mid g_i, g_j,...,g_k \in G\}$, where each $g$ is a gesture from $G$. Then each of these chains are considered as its own class, meaning that (in an example with 2 gesture long chains) a chain composed of gesture $A$ and then $B$ will be labelled as class $AB$, and a chain composed of gesture $B$ and then $A$ will be labelled as class $BA$. Therefore, to correctly identify a class, it is essential not only to recognise the individual gestures but also to understand the sequence in which they occur, making DVS-GC an action recognition task that demands perception of temporal order.

\begin{figure*}[]
\centering
\includegraphics[width=4.5in]{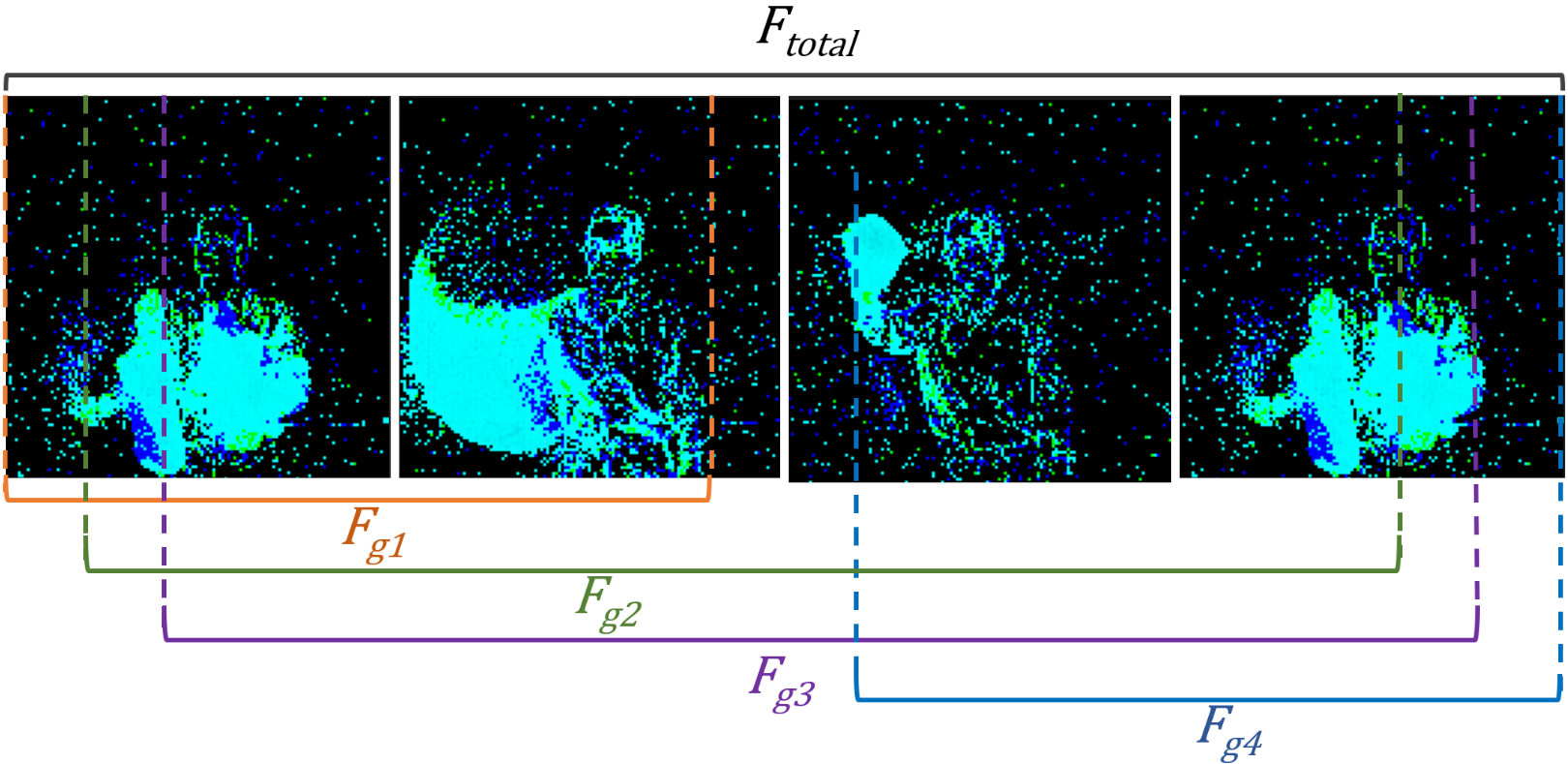}
\caption{Example gesture chain with variable $F_g$ duration ($\alpha_1=0.2$ and $\alpha_2=1$). The coloured underscores represent, for each gesture in the chain, the temporal window in which they could appear given the values of $\alpha_1$ and $\alpha_2$. This allows to understand why the gesture transition is not predictable and most time-steps have no guarantees of belonging to a certain position in the chain.}
\label{fig.gesture_chain}
\end{figure*}

\subsubsection{Event data processing}
When using neural networks to process streams of asynchronous events, it is common practice to discretise the time dimension by accumulating events in frames using a constant time window \cite{kaiser2020synaptic,zheng2020going, fang2021incorporating, vicente2021keys}. This allows to process the sequence with an arbitrary number of discrete time-steps and to train using methods such as Backpropagation Through Time (BPTT). 

Representing the event sequence as a function $e_{t,x,y,p}$, which has value $1$ when the position $(t_i,x_i,y_i,p_i)$ is active, and 0 otherwise, its discretised frame representation $F_{j,x,y,p}$ can be calculated as:

\begin{equation}
    F_{j,x,y,p}=\sum_{t=W(j-1)}^{Wj} e_{t,x,y,p}
\end{equation}
Where $j$ is the frame index (or time-step), $W$ the time window, $t$ represents time, $x$ and $y$ are the spatial coordinates and $p$ the polarity.
    

Naturally, the cost of this discretisation is that it makes it impossible to distinguish the precise timing or the relative order of occurrence of events within a frame.

We use this strategy for all our experiments, both with DVS-Gestures and DVS-GC. Then, at each time-step $j$ a new frame $F_j$ will be fed to the network.

When this quantisation is applied to the input of an SNN, the result is an approximation of the one we would obtain in an asynchronous implementation with infinite time resolution, as the effect is simply a reduction of time resolution, where groups of events are collapsed into the same time-stamp. In this scenario, the SNN's voltage leak can be represented as the percentage of voltage lost from one time-step to the next, meaning that this leak coefficient will be conditioned by the number of simulation time-steps. 

On the contrary, non-temporal ANNs cannot accumulate information from different time-steps, therefore collating events into frames is the only way in which they can combine information from events happening at different instants. This is further discussed in the results section.

\subsubsection{Creation of the action classes}
The creation of classes in DVS-GC is parametrized by the length of the gesture chain $L$ and the number of gestures $N$ used in the chain. Then the number of classes generated $C$ will be equal to all possible combinations (Eq. \ref{eq.repeat}). Alternatively, we also provide a class generation method which does not allow to repeat the same gesture in consecutive positions of the chain, reducing the possible permutations of the chain (Eq. \ref{eq.norepeat}). Using different values of $N$, $L$ and the methodology with and without repetition, we created different datasets that we use to evaluate our networks (results shown in section \ref{sec.results}).

\begin{equation}
    C=N^L\\\label{eq.repeat}
\end{equation}

\begin{equation}\label{eq.norepeat}
    C=N(N-1)^{L-1}
\end{equation}

\subsubsection{Chaining of events}

Given the stream of 4-dimensional events (x, y, time, polarity) provided by the DVS-Gesture dataset, as done in most state of the art systems \cite{kaiser2020synaptic,zheng2020going, fang2021incorporating, vicente2021keys}, we transform them into frames by accumulating events in a time window. 
The initial number of generated frames $F$ per sequence is user defined and will be constant for all instances in the dataset. The resulting frames have two channels, one for positive polarity and one for negative.

Given that the gesture instances are obtained from a set of users under different lighting conditions, the gesture chains are created combining the gestures from the same user under the same lighting condition. This avoids sudden changes in illumination or the appearance of the user, which could help the system identify the transition between a gesture and the next. It is also worth noting that the subjects are split in the Training and Testing sets following the original DVS-Gesture split, therefore, users appearing in the test set do not appear in the training set.

When building the gesture chains, having a constant number of frames for each gesture can also allow the machine to know when the transition will happen. To solve this, the duration in frames of each gesture $F_g$ is made variable. Let $F$ be the initial number of frames that the gesture sequences have and $F_{total}$ the target number of frames for the final gesture chains, which is user defined. Then, as seen in Eq. \ref{eq.fg}, the duration of each gesture $F_g$ will be a fraction of $F$ (parameterized by the coefficients $\alpha_1$ and $\alpha_2$) which satisfies that the total sum equal to $F_{total}$. 

\begin{equation}\label{eq.fg}
F_g  \in  [\alpha_1F, \alpha_2F] \mid \sum_{g=1}^L F_g = F_{total}
\end{equation}

Then, the value for $F_g$ is chosen randomly from the set we just defined, and the resampling from $F$ to $F_g$ is carried out by taking the first $F_g$ frames of the original sequence (Fig. \ref{fig.gesture_chain} demonstrates this variability visually). In our experiments, we define two datasets with $\alpha_1=0.5$ $\alpha_2=0.7$ and one with $\alpha_1=0.2$ $\alpha_2=1$. Both work well when using the sequences in DVS-Gesture because each gesture is repeated several times per recording, and therefore it is recognisable even after discarding a substantial part of the sequence.

Finally, when targeting a specific $F_{total}$, the number of initial frames $F$ that will allow the values of $F_g$ to have a uniform distribution between $\alpha_1F$ and $\alpha_2F$ is given by
Eq. \ref{eq.F}:

\begin{equation}\label{eq.F}
F = \frac{F_{total}}{L}\frac{2}{\alpha_1+\alpha_2}
\end{equation}

\subsection{Neural network architectures} \label{sec.NN_arch}

For our experiments, we make use of the state of the art SNN system presented in \cite{vicente2021keys}, the S-ResNet, which uses a LIF neuron model and reset by subtraction. The output layer is defined as a layer without leakage and without spiking activation, and its membrane potential after the last time-step provides the final class scores (complete specification in Appendix A). 
Apart from that, the S-ResNet uses BNTT as noramlization strategy. 

As seen in Eq. \ref{eq.bntt}, for a time-dependent input of $d$ dimensions $x_t = (x_{1,t}...x_{d,t})$, the method defines an individual BN module per time-step. This not only normalizes each feature $k$ (or convolutional channel in the case of CNNs) independently, as regular BN would do, but also defines independent statistics (mean $\mu_{k,t}$ and standard deviation $\sigma_{k,t}$) and learnable weights ($\gamma_{k,t}$ and $\beta_{k,t}$) per time-step $t$. 

\begin{equation}\label{eq.bntt}
   BNTT(x_{k,t}) = \gamma_{k,t} \frac{x_{k,t} -\mu_{k,t}}{ \sqrt{(\sigma_{k,t})^2+\epsilon}} + \beta_{k,t}
\end{equation}

In order to compare to non-spiking ANNs, we also define a non-spiking version of the same architecture. We substitute the neuron model by the Rectified Linear Unit (ReLU) activation function, and instead of BNTT we use regular BN. With these changes, the network becomes a conventional feed-forward ResNet. These networks process the input instantaneously, without temporal dynamics, which means that, for a sequence classification task such as action recognition, they can give an output per time-step but not a global one for the whole sequence. We solve this by adding the same output layer used by the SNN, which can be seen as a voting system that accumulates the outputs for all time-steps by summing them together. We refer to this network as ANN-BN. 

Additionally, in order to study the effect of the learnable weights in BNTT, we create a modified version of the non-temporal ANN that we call ANN-TW (ANN with temporal weight). This version adds a learnable weight $w_{l,t} \in \mathbb{R}^1$ per time-step $t$ at each layer $l$, which is used to scale the activation map after the convolution (Conv) and BN layers as: $y_{l,t} = BN(Conv(x_{l,t}))\cdot w_{l,t}$. As an alternative, we also define a version where each channel learns a different temporal weight. We refer to this network as ANN-TWC.

Finally, for the RNN vs SNN comparisons in Section \ref{sec.exp_rnn}, we define a different architecture with the objective of disentangling temporal processing from spatial processing. A non-spiking ResNet14 acts as spatial feature extractor, then (1 or 2) fully connected "temporal layers" of 128 features are appended before the final classification layer, to act as temporal feature extractor. As temporal layers we test feed-forward SNN layers, recurrent SNN layers (RSNN), vanilla RNN and LSTM. 

\section{Results}\label{sec.results}

In this section, we first prove how a network without the capability for temporal feature extraction (ANN-BN) can solve the classification task in DVS-Gesture but fails to do so in the new DVS-GC, which demands a perception of temporal order. We then demonstrate how, in contrast, an SNN of the same architecture learns to perceive the temporal dependencies in DVS-GC. 
From there, we evaluate the effects of the membrane potential reset strategy, voltage leak, time-dependent weights, and time-dependent normalization. Finally, we compare SNNs to conventional RNNs.

\subsection{DVS-Gesture evaluation}

Many previous works reporting accuracy performances in DVS-Gestures have taken the approach of training with the whole training set, evaluating test set performance through training epochs, and then reporting the highest test accuracy as the final test accuracy. We consider this approach to be reporting validation accuracy rather than test. Therefore, in our setup, we only evaluate the test set after the training is complete, without using its value to tune the training.

Table \ref{tab.gestures} shows how both SNN and ANN achieve high accuracy in the DVS-Gestures task. As previously stated, the ANN final prediction is just a sum of the individual predictions made at each time-step. Each of these is made using the information from a frame which integrates the events received within a time window; in the case of our experiments, the time window is $\frac{1}{50}$ of the total event sequence. 

The ANN has no way of combining information from different frames and has no notion of the timing in which they occurred, hence, it does not perceive the timing or relative order of the events. Still, we cannot say that the features it uses are strictly non-temporal. When accumulating these events into frames, only the ones which are close in time will be integrated into the same frame, meaning that the spatial features the network will calculate are still dependent on event timing. This makes the appropriate wording a sensible matter: The ANN does not implement working memory neither does it implement temporal feature extraction, still, the spatial features it extracts in this scenario are dependent on event timing, therefore, given that a temporal feature is any attribute of the data that is explicitly related to time, these features can be considered a type of temporal feature.
This explains why a network without temporal feature extraction can solve the DVS-Gestures task, and how solving this task does not require to perceive the temporal ordering of events, but only to integrate events which are close in time so that spatial features become apparent. Then, a system designed for the classification of static images can perform the task.

For completeness, we also report the accuracy obtained by the SNN with conventional BN (SNN-BN) and the accuracy of the ANN with BNTT (ANN-BNTT). It can be seen how the ANN does not benefit from the time-dependent computations of BNTT and obtains a very similar result. On the contrary, the SNN performance decreases when using regular BN, demonstrating how, for a system where activity statistics change through time such as SNN, timing-aware normalization is beneficial. 

\begin{table}[]\centering 
\caption{\label{tab.gestures}  Test performance on DVS-Gesture. SNN* was initialized with pre-trained weights as proposed in \cite{vicente2021keys}. Training and testing were run for 3 times, accuracies presented as mean ± std.}
\begin{adjustbox}{max width=0.5\textwidth}
\begin{tabular}{lcc}
\hline
\textbf{Network} & \textbf{Normalization} & \textbf{DVS-Gesture Accuracy} \\ \hline
SNN & BN    & 70.31 ± 3.27 \%   \\ 
SNN & BNTT  & 89.82 ± 1.50\%\\ 
SNN* & BNTT  & 94.84 ± 1.06 \% \\ 
ANN & BN      & 97.35 ± 0.45 \%    \\ 
ANN & BNTT & 96.95 ± 0.61\%    \\ \hline
\end{tabular}
\end{adjustbox}

\end{table}

\subsection{DVS-Gesture-Chain evaluation}\label{sec.dvs-gc_ex}
Using the methodology described in the previous section \ref{sec.dvs-gc} we created three DVS-GC datasets, which are summarised in Table \ref{tab.datasets}. Datasets \textit{81-p} and \textit{96-p} define a smaller variability for the duration $F_g$ of each individual gesture ($\alpha_1 = 0.5$, $\alpha_2 = 0.7$), while \textit{96-u} defines a larger one, making it much harder to predict the transition between gestures in time (Fig. \ref{fig.gesture_chain} demonstrates this variability visually). 

For all three, we create a validation set with 20\% of the training data and evaluate it at every epoch. The test performance is then evaluated using the weights with the highest validation accuracy.

\begin{table}[h!]\centering 
\caption{\label{tab.datasets} Parameters per dataset. In the naming convention, \textbf{-p} stands for predictable time windows while \textbf{-u} stands for unpredictable time windows. }
\begin{adjustbox}{max width=0.5\textwidth}

\begin{tabular}{ccccccc}
\hline
\textbf{Name} & \textbf{N} & \textbf{L} & $\boldsymbol{\alpha_1}$ & $\boldsymbol{\alpha_2}$ & \textbf{Repetition} & \textbf{\# classes} \\ \hline
\textbf{81-p} & 3 & 4 & 0.5 & 0.7 & Yes (Eq. \ref{eq.repeat}) & 81 \\ 
\textbf{96-p} & 3 & 6 & 0.5 & 0.7 & No (Eq. \ref{eq.norepeat}) & 96\\ 
\textbf{96-u} & 3 & 6 & 0.2 & 1 & No (Eq. \ref{eq.norepeat}) & 96\\ \hline
\end{tabular}
\end{adjustbox}

\end{table}

\subsubsection{ANN vs SNN and time-dependent weights}\label{sec.ANN_vs_SNN}

We begin by evaluating the networks on the \textit{81-p} and \textit{96-p} datasets. As seen in Table \ref{tab.dvsgh}, now that the task requires distinguishing the ordering of the events, the non-temporal ANN (ANN-BN) fails to solve it. Moreover, its accuracy value implicitly reveals the computations performed by the network: Taking the 81 class data set as an example, one can see how the ANN accuracy (16.91\%±0.5) is higher than random chance (1.23\%). This is because the network is capable of detecting the gestures present in the sequence and the number of times they appear, but is unable to perceive their ordering. With such conditions, and assuming a perfect accuracy in gesture detection, the probability of correctly classifying a sequence for the 81 class dataset is $p_d = 16.05\%$ (proof in Appendix B).

\begin{table}[]\centering 
\caption{\label{tab.dvsgh} Test performance on
DVS-GC. Training and testing were run for 3 times, accuracies presented as mean ± std. }
\begin{adjustbox}{max width=0.5\textwidth}
\begin{tabular}{lccl}
\hline
\textbf{Network} & \textbf{Normalization}  & \textbf{Dataset}  & \textbf{Accuracy} \\ \hline
ANN        & BN & 81-p & 16.91 ± 0.50 \%   \\ 
ANN & BNTT &81-p  & 99.52 ± 0.31 \% \\  
ANN & BN + TW &81-p & 89.44 ± 5.74 \%   \\ 
ANN & BN + TWC &81-p & 91.08 ± 8.10 \%    \\
SNN        & BN & 81-p &  86.00 ± 2.38 \% \\ 
SNN & BNTT &81-p & 95.83 ± 0.62 \%   \\ 
 \hline

ANN        & BN & 96-p & 12.96 ± 0.74 \%   \\ 
ANN & BNTT &96-p  & 99.52 ± 0.55 \% \\ 
SNN        & BN & 96-p &  80.62 ± 2.76 \% \\ 
SNN & BNTT &96-p & 96.32 ± 0.02 \%   \\ \hline

ANN  & BNTT & 96-u & 74.66 ± 0.64 \%   \\ 
SNN  & BNTT & 96-u & 91.16 ± 1.30 \%   \\ \hline

\end{tabular}
\end{adjustbox}
\end{table}

On the other hand, the results show how the SNN still achieves high accuracy on these same datasets, implying that its temporal dynamics allow to perceive order in time. In order to analyse which components of the network enable this capacity, we also test the performance of the SNN with conventional BN (SNN-BN) and the accuracy of the ANN with BNTT (ANN-BNTT). The accuracies obtained by both systems indicate that they are successfully learning to perceive order, meaning that both, spiking neurons and BNTT can enable a neural network to recognise temporal sequences on their own. Additionally, it is also worth noticing how, for the \textit{81-p} and \textit{96-p} datasets, the ANN with BNTT is more accurate than the SNN.

In BNTT, the perception of order is gained by learning time-dependent values that are used to scale the activation maps of the network, providing temporal attention. In order to decorrelate this capacity from the normalization strategy, we create two modified versions of the non-temporal ANN with regular BN which implement learnable temporal weights, ANN-TW and ANN-TWC (introduced in Section \ref{sec.NN_arch}).

The performance of ANN-TW (Table \ref{tab.dvsgh}) proves how a single time-dependent weight per layer is enough to recognise the temporal sequences in \textit{81-p} and how the learned value does not need to be different between channels for temporal perception purposes. Still, the ANN-TWC obtains a slightly higher accuracy. Apart from that, the performances of both networks are lower to those of the systems using time-dependent normalization statistics, proving how these are not essential but indeed beneficial. A BNTT ablation study is available in Appendix C.

Finally, we evaluate the performances of the networks with the \textit{96-u} configuration, where the variability of the duration $F_g$ of each individual gesture is higher.
The results (last two rows of Table \ref{tab.dvsgh}) demonstrate how this set-up greatly decreases the performance of the ANN-BNTT, meaning that temporal attention is not enough for the task. In contrast, SNN-BNTT exhibits a smaller decrease and still solves the task with high accuracy, proving how the capacity of SNNs for spatio-temporal feature extraction goes beyond that of temporal attention. The complete analysis justifying these results is provided in Section \ref{sec.analysis}.

\subsubsection{Leak and reset mechanism}

When using regular BN, the SNN does not have temporal weights, making voltage leak the only time-aware component in the network. This motivates us to explore its relevance to solving the task.

Table \ref{tab.leak} compares the results of the same network trained with LIF neurons and IF neurons (no leak). Because the performance comparison between LIF and IF can be affected by the chosen leak coefficient, we searched for its optimal value through hyper-parameter search. We find the best results with $0.87$ for the 81 class dataset and $0.80$ for the 96 class one. 

The original network, which uses reset by subtraction, suffers a major performance drop when not equipped with leak. We found that the reason behind this is the excess of voltage in neurons reset by subtraction, which can trigger delayed spikes that slow down adaptation to newer inputs. We quantify this effect by means of what we call "repetition error" (R-error). The R-error is measured as the percentage of wrong classifications where at least one of the miss-classified gestures in the chain has been predicted to be the same as the preceding one.  As there are three different gestures to choose from at each position in the chain, the standard R-error is 33.3\%. Values higher than this one will indicate a tendency towards repeating previous predictions. (Notice that the 96-class dataset does not allow repetition in its classes and therefore cannot present R-error, still, not clearing old voltage also decreases the performance in it)

These results demonstrate how voltage leak prevents old information from corrupting current calculations and solves the voltage stagnation problem caused by the reset by subtraction. In addition to that, we evaluated how a reset to zero strategy can also prevent this same issue (Table \ref{tab.leak}), as it does not retain any voltage after spiking, hence not generating delayed spikes. 

Interestingly, reset to zero consistently achieves the best performance when paired with IF neurons, while implementing leak decreases its accuracy. Not implementing voltage leak will mean that neurons close to reaching the spiking threshold will remain in that state, even after the stimulus that was triggering them is long finished. This will make them prone to spiking prematurely in later processing, arguably causing noisier computations. On the other hand, leakage represents the progressive loss of the short-term memory of the network, which also has the potential to disrupt computations. The fact that leak is not beneficial for the task at hand might indicate that the former issue is not prevalent. One possible explanation can be that, when a new gesture comes, initial noise in the spiking pattern is still superseded by later detections due to data redundancy through time (the gesture can be continuously detected throughout a window of time).

\begin{table}[]\centering 
\caption{\label{tab.leak} Test accuracy and R-error of the SNN-BN under different setups. IF neurons do not leak. Zero stands for reset to zero and Sub for subtraction. LIF neurons use a leakage factor of 0.87 except for LIF-Sub in 96-u, which uses 0.80. Training and testing were run for 3 times, accuracies presented as mean ± std.}
\begin{adjustbox}{max width=0.5\textwidth}
\begin{tabular}{ccccc}
\hline
\textbf{Neuron} & \textbf{Reset} & \textbf{Dataset} & \textbf{Accuracy} & \textbf{R-error} \\ \hline
LIF & Sub & 81-p & 86 ± 2.38 \% & 33.54 ± 5.20 \%   \\ 
IF & Sub & 81-p & 48.27 ± 1.38 \% & 70.41 ± 2.44 \%  \\ 
LIF & Zero & 81-p & 82.31 ± 2.14 \% & 35.30 ± 8.09 \%   \\ 
IF & Zero & 81-p & 92.59 ± 2.67 \% & 31.01 ± 2.15 \%  \\ \hline 
LIF & Sub & 96-u & 68.74 ± 1.45 \% & n/a   \\ 
LIF & Zero & 96-u & 63.58 ± 3.09 \% & n/a  \\ 
IF & Zero & 96-u & 71.40 ± 4.55 \% & n/a  \\ \hline

\end{tabular}
\end{adjustbox}
\end{table}

\subsubsection{RNN vs SNN}\label{sec.exp_rnn} 

After demonstrating how SNNs can perform temporal computations without the need for recurrent connections, we further validate our results by comparing their performance with that of RNNs. For this, as introduced in Section \ref{sec.NN_arch}, we define a different architecture based on a non-spiking ResNet14, which acts as spatial feature extractor, and append (1 or 2) fully connected "temporal layers" before the final classification layer, to act as temporal feature extractor. We test feed-forward SNN layers, recurrent SNN layers (RSNN), vanilla RNN and LSTM as temporal layers.

Table \ref{tab.rnn} shows how, for \textit{96-u}, SNNs outperform vanilla RNNs while LSTMs outperform SNNs. The RSNN demonstrates a substantial improvement with respect to the SNN when using two layers, getting closer to the LSTM performance. On the other hand, in the dataset with predictable time windows, \textit{81-p}, all networks perform at a similar level, implying that all networks manage to exploit its predictable time windows, arguably, demonstrating time-dependent feature extraction.

The performance differences between SNN, RSNN, RNN, and LSTM are well justified by their computing principles, which we analyse in section \ref{sec.analysis_rnn}.

\begin{table}[]\centering 
\caption{\label{tab.rnn} Test accuracy in DVS-GC \textit{81-p} and \textit{96-u}. SNNs use IF neurons and reset to zero. Training and testing were run for 3 times, accuracies presented as mean ± std. "\# TL" stands for number of temporal layers."\# params" presents the number of parameters used in the temporal layers as a factor of the parameters of a 128-dimensional dense layer. *RNN presents the maximum accuracy instead of the mean, as numerous trials failed to learn.}
\begin{adjustbox}{max width=0.5\textwidth}
\begin{tabular}{ccccc}
\hline
\textbf{Temporal layers} & \textbf{\# TL} & \textbf{\# params} & \textbf{Dataset} & \textbf{Accuracy} \\ \hline
SNN &1& $\times$1 & 96-u& 61.53 ± 3.27 \% \\
SNN &2& $\times$2 & 96-u& 67.09 ± 1.57 \% \\
RSNN &1& $\times$2 & 96-u& 64.62 ± 0.69 \% \\
RSNN &2& $\times$4 & 96-u& 77.80 ± 1.92 \% \\
RNN &1& $\times$2 & 96-u &  44.08* \% \\
RNN &2& $\times$4 & 96-u &  65.58* \% \\
LSTM &1& $\times$8 & 96-u& 87.31 ± 2.79 \% \\
LSTM &2& $\times$16 & 96-u& 88.80 ± 3.47 \% \\\hline

SNN & 2& $\times$2 & 81-p& 84.60 ± 3.08 \% \\
RSNN & 2& $\times$4 & 81-p& 86.47 ± 1.44 \% \\
RNN & 2& $\times$4 & 81-p & 72.32 ± 16.91 \% \\
LSTM & 2& $\times$16 & 81-p& 89.83 ± 3.34 \% \\
 \hline

\end{tabular}
\end{adjustbox}
\end{table}

\section{Analysis of temporal computations}\label{sec.analysis}
After proving through empirical results how spiking neurons and time-dependent weights enable temporal order perception, in this section we analyse the in-depth mechanics that implement this capability.
\subsection{Temporal attention analysis}
Networks with time-dependent weights such as ANN-TW or those using BNTT use temporal attention to store the time at which a visual detection occurred. This is achieved by constraining the activation of certain layers or channels to a time window, then the feature detected by those neurons will be known to happen within that time-window.

\begin{figure}[]
\centering
\includegraphics[width=0.5\textwidth]{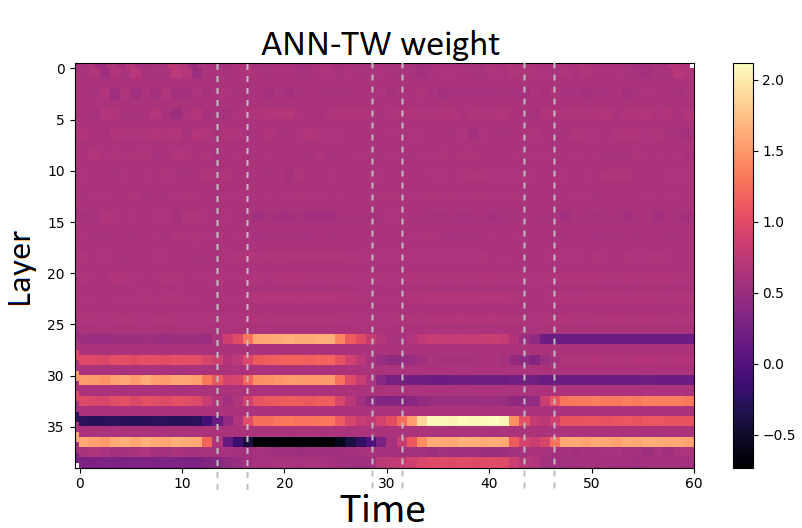}
\caption{Value of the time weight in TW-ANN. Dotted lines highlight the gesture transition zone. The last two layers of the graph correspond to the layers in the residual connection downsampling. Trained in the \textit{81-p} DVS-GC.}
\label{fig.ann-tw}
\end{figure}

In order to prove how the networks are using this strategy, in Fig. \ref{fig.ann-tw}, we visualize the value of the temporal weight in ANN-TW when trained in the \textit{81-p} dataset. Notice that, when designing DVS-GC, we made the gesture chaining procedure variable in time, so that the transition between gestures does not always happen in the same time-step. Now, visualising the graph, it can be seen how the network learned to reduce the weight in the uncertainty zone of the transition and defined its detection time windows between the time-steps which are guaranteed to belong to the \textit{n-th} gesture. Then, we observe how the weight restricts the last layers to only be active in time windows corresponding to specific positions in the 4-gesture chain. This specializes different layers in detecting gestures at certain positions in the chain, acting as a temporal attention coefficient. This is equivalent to associating timestamps to the detected spatial features and then combining this information in the last layer by accumulation. This last layer is the only element in the system implementing memory for the non-spiking networks. 
The same principle was proven for BNTT in Appendix D.
 
Given this computational logic, it is then clear why in the \textit{96-u} dataset the performance of these networks dropped. With $\alpha_1=0.2$ and $\alpha_2=1$ time-steps are not guaranteed to contain a specific position in the chain (except for the first and last gestures), since the transition zones now overlap. Therefore, in that scenario, time-dependent features calculated using temporal attention are not a reliable descriptor. 

\subsection{Spiking neuron analysis}
Unlike temporal attention, spiking neurons achieve high accuracy in all three datasets. This shows how SNN can perform two types of spatio-temporal tasks:

\begin{enumerate}
    \item Sequence recognition with predictable action time windows in the \textit{81-p} dataset.\label{t1}
    \item Sequence recognition with unpredictable time windows in the \textit{96-u} dataset.\label{t2}
\end{enumerate}
Task \ref{t1} can be solved by means of time-dependent features, as shown by the analysis performed on temporal weights. Moreover, as this dataset allows repetition, these kind of features are indispensable in order to distinguish individual gestures when the same one is repeated in succession. 

Task \ref{t2}, to the best of our knowledge, can only be solved by recognising each gesture transition in the chain, as the timing of an action is not enough to find its chain location and, therefore, the relative order of appearance is the only usable information. 

Following that logic, by performing successfully in Task \ref{t2}, SNNs demonstrate that they can detect gesture transitions in a time-invariant manner, while in Task \ref{t1}, they demonstrate the ability to localise gestures in time. Such behaviour implies that spiking neurons enable time-invariant spatio-temporal feature extraction as well as time-dependent feature extraction.

\begin{figure*}[]
\centering
\includegraphics[width=0.95\textwidth,trim={0in 9in 0in 0in},clip]{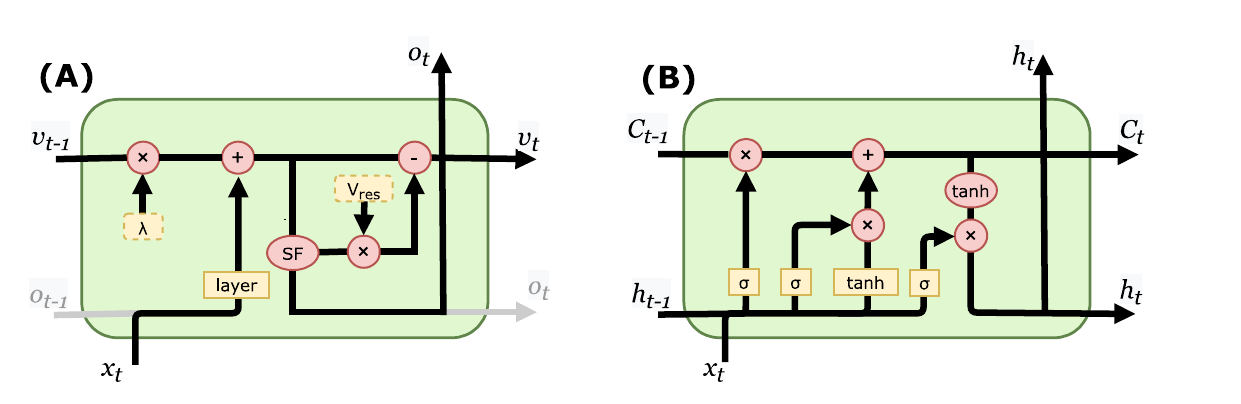}
\caption{\label{fig.lif} (A): Diagram of a layer of LIF neurons. $layer$ are the synaptic weights, $SF$ the spiking function, $V_{res}$ the voltage reset value. Gray lines show the architecture with recurrent connections, without them, the architecture is feed-forward. (B): LSTM diagram. $C_t$ is the cell state, $h_t$ the hidden state and output, the yellow $tanh$ is a layer of synaptic weights with Hyperbolic Tangent activation. $\sigma$ stands for the gating layer with Sigmoid activation.}
\end{figure*}
\subsubsection{Comparing SNN to RNN}\label{sec.analysis_rnn}
A layer of spiking neurons can be seen as a recurrent cell where the current input $x_t$ passes through a layer and is summed to the previous state $v_{t-1}$. The contribution of $v_{t-1}$ is weighted by the leak factor, and the final voltage $v_t$ is decreased by subtracting voltage in the event of a spike. Therefore, they retain memory by integrating inputs over time until the threshold is surpassed. Then, when the information is released in form of a spike, it is deleted from memory through the reset mechanism. On the contrary, in a non-spiking vanilla RNN, the neuron itself has no memory, recurrency is implemented only at layer level by defining a recurrent synapse from each neuron to itself and lateral synapses within the layer. Therefore, it is to be expected for SNN/RSNN to have better performance than RNNs.

On the other hand, LSTM adds a cell state to their computations to integrate inputs through time, like spiking neurons do, but, as seen in Figure \ref{fig.lif}, it has three differences: First, the integration is weighted by two gating layers, the input and forget gates, while LIF neurons compute which information to forget through the reset mechanism and the leak factor. Second, in LSTM the non-linearity is applied before integration, while the spiking neuron applies its non-linearity (thresholding) only to the output. Finally, in an LSTM the output is weighted by another gating layer.This side by side comparison illustrates how spiking neurons define a computing principle similar to LSTM units, but without the use of gating layers, resulting in a lighter network. Additionally, it allows to understand how, as seen in the experiments, an internal state can be enough to calculate temporal features. Recurrent connections can be beneficial, but they are not indispensable.

\section{Discussion}

In this work, we showed how spiking neurons can be exploited to solve temporal tasks without the need of recurrent synapses. This proves how their temporal dynamics are not only a vehicle for computational efficiency, but also a tool for the extraction of temporal features. This can allow to bypass the need for recurrent connections when a lighter network is needed and to reuse feed-forward networks for temporal tasks. Moreover, the parallelism drawn between LSTM and SNN allows to appreciate how SNN computation is closer to LSTM than to vanilla RNNs. Understanding their similarities and differences allows to make informed choices when designing temporal processing systems and paves the way to distilling more biologically inspired principles into machine learning. Additionally, it also contributes to closing the gap between neuroscience and machine learning knowledge.

In our experiments, we evaluated the two components that allow an SNN to clean its memory, the leak and reset mechanism. The effect of the leak factor has been previously evaluated in static data \cite{CHOWDHURY202183}, but understanding its relevance for temporal computations was still necessary. Our results contribute to develop this understanding by showing how voltage leak prevents old information from stagnating in the network when using reset by subtraction. Looking at the reset strategy, we find that zero-reset also solves the aforementioned problem. Reset by subtraction has been a popular option given that it prevents loss of information, and has been proved especially useful in ANN to SNN conversion approaches \cite{10.3389/fnins.2017.00682}. Still, our results indicate that retaining such information can come at the cost of slower adaptation to dynamic inputs. Therefore, we believe that this effect should be taken into account when designing SNNs for temporal processing tasks, and appropriately handling it will lead to improved results, as shown in our experiments.

Additionally, the analysis of temporal weights demonstrated a clear use case for temporal attention, showing how time-dependent features are learned by a network when the meaning of the events is dictated by their timing. In this work, the implementation of temporal weights and time-dependent normalization requires learning a parameter per time-step, which would be a limitation for inputs of variable length. Still, it is enough to demonstrate the aforementioned computing principle. Moreover, it serves as a tool to prove which tasks can be solved with time-dependent features without the need for time-invariant ones. 

These insights were obtained thanks to the newly proposed DVS-GC, a task which was created by means of a novel chaining technique. The relevance of this task is that, first, it fulfils the current need for event-based action recognition datasets. Apart from that, it provides an approach that allows the creation of controlled scenarios in order to evaluate specific capacities of a learning system. The datasets built in this work serve as examples, where \textit{81-p} and \textit{96-p} could be solved by timing-aware features, and \textit{96-u} could only be solved with time-invariant spatio-temporal features. Moreover, the chains can be made arbitrarily long, which allows to test the limits of a system's memory. Looking ahead, the results provided in the proposed DVS-GC configurations can serve as a baseline when evaluating new systems. Apart from that, if a more challenging task is needed, the method allows to build longer sequences with more gestures.

\section*{Acknowledgement}
This work was supported by the US Air Force Office of Scientific Research under Grant for project FA8655-20-1-7037. The contents were approved for public release under case AFRL-2023-1422 and they represent the views of only the authors and does not represent any views or positions of the Air Force Research Laboratory, US Department of Defense, or US Government.

\section*{Appendix}

\appendix

\section{Neural Network architecture}\label{app.net}
Here we present the full specifications of the implemented neural network and the neuron model.

The neurons in the SNN are defined by the LIF neuron model. Let $i$ be a post-synaptic neuron, $u_{i,t}$ its membrane potential, $o_{i,t}$ its spiking activation and $\lambda$ the leak factor (which we set to 0.874 following the original S-ResNet paper, unless specified otherwise). The index $j$ represents the pre-synaptic neuron and the weights $w_{i,j}$ dictate the value of the synapses between neurons. Then, the iterative update of the neuron activation is calculated as follows:
\begin{equation}\label{eq.spike}
    o_{i,t} = g \left( \sum_{j} (w_{i,j}o_{j,t})\ + \lambda \cdot u_{i,t-1} \right)
\end{equation}
where $g(x)$ is the thresholding function, which converts voltage to spikes:

\begin{equation}\label{eq.threshold}
g(x) = \begin{cases}
    & \text{1, if } x \geq  U_{th} \\ 
    & \text{0, if } x < U_{th} 
\end{cases}
\end{equation}
After spiking, a reset is performed by the subtraction $u_{i,t}^*=u_{i,t} - U_{th}$, where $u_{i,t}^*$ is the membrane potential after resetting. In experiments using zero reset, $u_{i,t}^*=0$.

We use a Spikes to Spikes (S2S) implementation for the residual connection and the output layer is defined as a layer without leakage ($\lambda = 1$) and without spiking activation ($U_{th}=\infty$). This output layer accumulates the network output through all time-steps and its voltage $u_{i,t}$ at the last time-step $t=T$ provides the final class scores (Eq. \ref{eq.voting}). At training time, these are compared to the ground truth labels by means of a cross-entropy loss and the network is trained by Backpropagation Through Time (BPTT), using Stochastic Gradient Descent with a momentum of 0.9.

\begin{equation}\label{eq.voting}
    u_{i,T} =  \sum_{t}^T\sum_{j} (w_{i,j}o_{j,t})
\end{equation}

Given the non-differentiability of the thresholding function, a triangle shaped surrogate gradient is used as its derivative (Eq. \ref{eq.triangle}). We use $\alpha=0.3$.
\begin{equation}\label{eq.triangle}
    \frac{\partial o_{t,i}}{\partial u_{t,i}} = \alpha \max \{0, 1-\vert u_{t,i}\vert\}
\end{equation}

We use a depth of 38 layers and a width of 32 base filters unless otherwise specified. Training was performed using 4 Nvidia GeForce GTX 1080 Ti GPUs.

\section{Proof for the probability of correct classification without order perception}\label{app.proof}
Assume a system with perfect gesture classification but no perception of order that is evaluated in DVS-GC. This system will know which gestures are present in the sequence and the number of times they appear, but will be unable to perceive their ordering.

For this system, depending on the number of detected gestures, the candidates to be the correct output are reduced. Therefore, to calculate its accuracy, we can calculate the probability of correctly classifying each individual class $x_i$ in the dataset and then, assuming a constant number of class examples, their average will be the final accuracy.

We calculate this for the 81-class DVS-GC dataset in equation \ref{eq.prob}.

\begin{gather}
p(x) = \begin{cases}
    & 1 \text{, if gesture repeated 4 times } \\
    & \frac{1}{4} \text{, if gesture repeated 3 times } \\
    & \frac{1}{6} \text{, if 2 gestures repeated 2 times } \\
    & \frac{1}{36} \text{, otherwise }
\end{cases}
\nonumber\\
P_d = \frac{1}{C}\sum_i^C p(x_i)= 
\nonumber\\\label{eq.prob}
=\frac{1}{81}(3\cdot 1 + 24\cdot \frac{1}{4}
+ 18\cdot \frac{1}{6} + 36\cdot \frac{1}{36}) = 0.1605
\end{gather}

\section{BNTT Ablation study} \label{app.bntt_abl}
The BNTT module has 4 time-varying parameters, namely mean, variance, $\gamma$ weight and $\beta$ weight. In order to analyse their role in temporal understanding, we perform an ablation study where we eliminate the temporal dimension of some of this components by averaging across all time-steps. This allows to isolate the temporal performance of individual components and evaluate the accuracy degradation. Note that the experiment is performed by first training the regular network and then averaging the necessary parameters, with no retraining after the ablation.

Table \ref{tab.bntt} presents the results for each of the 4 BNTT parameters in isolation (all the other parameters were averaged in time). It can be seen how any of them is enough to maintain accuracy well above 16.05\%, meaning that they all encode part of the temporal attention learned by the network. Still, there is a clear difference in accuracy between them, with $\gamma$ and $\beta$ weights having the highest accuracy and the mean having the lowest one. Of course, after averaging all of them, the network falls back to the performances seen by the ANN with regular BN.

\begin{table}[h!]\centering 
\caption{\label{tab.bntt} Test accuracy of the ANN-BNTT after averaging certain components through time. "Non-averaged components" indicates which parts have not been averaged and therefore are still time-dependent.}
\begin{tabular}{lc}
\hline
\textbf{Non-averaged components}  & \textbf{Accuracy} \\ \hline
Full BNTT  & 99.16 \%   \\ 
$\beta$ weight & 97.63 \% \\ 
$\gamma$ weight & 93.77 \% \\ 
Variance & 72.24 \% \\ 
Mean & 35.96 \% \\ 
None & 17.09 \% \\ \hline

\end{tabular}

\end{table}

\section{BNTT's temporal attention}\label{app.bntt_att}

\begin{figure*}[h!]
\centering
\includegraphics[width=\textwidth]{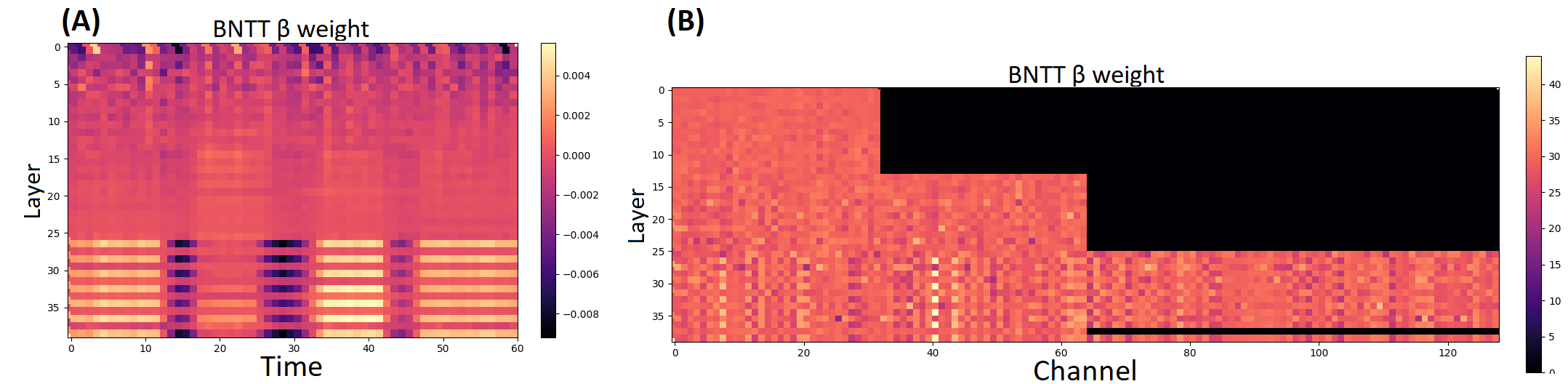}
\caption{(A): Bias weight average value across channels in the BNTT layers of ANN-BNTT. (B): Value of the center of mass in the time dimension (60 time-steps) of the bias weight of the BNTT layers in ANN-BNTT. The last two layers of all graphs correspond to the layers in the residual connection downsampling. Trained in the 81-class DVS-GC.}
\label{fig.bntt_att}
\end{figure*}

Following on the study presented in section 4.2.1
, we analyse the temporal attention in BNTT. Fig. \ref{fig.bntt_att}.A displays the value of BNTT's $\beta$ weight, which scales neuron activations. Unlike the ANN-TW graph, this one does not show large changes in the coefficient value. This is because the $\beta$ weight has a different value per channel, something that is not visible in Fig. \ref{fig.bntt_att}.A, as it averages through channels. Therefore, in order to visualize the temporal windowing of the BNTT weight, in Fig.  \ref{fig.bntt_att}.B we plot the center of mass $m$ in the time dimension for the weights at each channel as:
\begin{equation}
    m = \frac{1}{T}\sum_t^T (x_t-min(x)) t 
\end{equation}
Where $x$ is a vector that contains a weight value $x_t$ per time-step $t$. 

With a uniform distribution of weight through the time-steps, the center of mass would have a value equal to $T/2$, which in the case of our network would be 30. Consequently, all the values in Fig. \ref{fig.bntt_att}.B that are far from this number are indicators of the existence of a time window. 
It can be seen how the centre of mass varies among different channels, demonstrating how they specialise on detecting features inside different temporal windows. This proves how BNTT hard-codes a different temporal attention for each channel in a layer.



\bibliography{bib_collection_28-04-22}
\bibliographystyle{unsrt}
\end{document}